\documentclass[conference]{IEEEtran}
\IEEEoverridecommandlockouts
\usepackage{cite}
\usepackage{amsmath,amssymb,amsfonts}
\usepackage{booktabs}
\usepackage{float}
\usepackage{placeins}
\usepackage{graphicx}
\usepackage{textcomp}
\usepackage{afterpage}
\usepackage{xcolor}
\usepackage{algorithm}
\usepackage{algpseudocode}

\usepackage[colorlinks=true]{hyperref}
\def\BibTeX{{\rm B\kern-.05em{\sc i\kern-.025em b}\kern-.08em
		T\kern-.1667em\lower.7ex\hbox{E}\kern-.125emX}}

\makeatletter
\let\old@ps@headings\ps@headings
\let\old@ps@IEEEtitlepagestyle\ps@IEEEtitlepagestyle
\def\confheader#1{%

\def\ps@IEEEtitlepagestyle{
\old@ps@IEEEtitlepagestyle
\def\@oddhead{\strut\hfill#1\hfill\strut}
\def\@evenhead{\strut\hfill#1\hfill\strut}
}
\ps@headings
}
\makeatother
\confheader{
\small{Proceedings of the 13th RSI International Conference on Robotics and Mechatronics (ICRoM 2025), Dec. 16-18, 2025, Tehran, Iran} 
}

\usepackage[pscoord]{eso-pic}
\newcommand{\placetextbox}[3]{
\setbox0=\hbox{#3}
\AddToShipoutPictureFG*{ \put(\LenToUnit{#1\paperwidth},\LenToUnit{#2\paperheight}){\vtop{{\null}\makebox[0pt][c]{#3}}}
}
}
\placetextbox{.5}{0.055}{\textbf{\small{Proceedings of the 13th RSI International Conference on Robotics and Mechatronics (ICRoM 2025), Dec. 16-18, 2025, Tehran, Iran}}}

\begin{document}

\title{Iterative Grasp Pose Refinement: A Deep Reinforcement Learning Approach for 2D Vision}

\author{
	Amir Arsalan Nematollahi$^{\ast}$, Shayan Ahmadi$^{\ast}$, Mehdi Tale Masouleh, Ahmad Kalhor \\
	\textit{Human and Robot Interaction Laboratory, School of Electrical and Computer Engineering,} \\
	\textit{University of Tehran, Tehran, Iran} \\
	\{a.nematollahi, shayanahmadi, m.t.masouleh, akalhor\}@ut.ac.ir
	\thanks{$^{\ast}$Equal contribution.} 
}

\maketitle

\begin{abstract}
Developing robots capable of understanding and manipulating objects requires compact, interpretable, and generalizable representations. This work proposes a reinforcement learning-based framework for robotic grasp refinement, integrating keypoint-based object representations with a Deep Q-Network (DQN). Using 2D overhead images captured in a simulated environment, a geometric-based algorithm generates initial grasp candidates, which are iteratively refined by the proposed framework, transforming failed grasps into successful ones. Experiments conducted on 300 objects from the Dex-Net dataset using a UR5 manipulator demonstrate the framework’s effectiveness, achieving a 100\% success rate on objects previously deemed ungraspable by geometrical methods. The framework's sim-to-real transferability is further validated through physical experiments on a Delta parallel robot, where a refined grasp successfully manipulates an object that was previously ungraspable. The findings underscore the effectiveness of reinforcement learning in addressing challenges in robotic grasping, offering a scalable and adaptable solution for contact-rich manipulation tasks.
\end{abstract}

\begin{IEEEkeywords}
	Robotics, Grasping, Reinforcement Learning, Grasp Refinement, Object Grasping
\end{IEEEkeywords}

\section{Introduction}
Object manipulation remains a fundamental challenge in robotics, with crucial applications in industrial automation and domestic assistance. Despite significant progress, current systems still lack the human-like dexterity and adaptability needed for unstructured environments. To bridge this gap, data-driven approaches like deep Reinforcement Learning (RL) have become powerful tools. By leveraging trial-and-error interaction, deep RL learns robust and adaptable grasping strategies directly from sensory data~\cite{sutton2018reinforcement, franccois2018introduction, sekkat2024review}.

Effective object representations are crucial for enhancing these learned behaviors. Keypoint representations, in particular, allow a robot to focus on critical geometric features, facilitating more precise and efficient manipulation. Integrating keypoint detection with RL is a promising path toward creating reliable, versatile grasping systems that can generalize across diverse objects~\cite{manuelli2019kpam, fang2020learning, koppula2013learning}.

Data-driven grasping pipelines primarily rely on supervised learning or reinforcement learning~\cite{dukor2021survey}. Supervised approaches train on large, pre-labeled datasets of grasp candidates, a method that can be computationally demanding despite yielding strong generalization~\cite{satish2019policy, hosseini2020improving, mahler2017dex}. RL has proven effective for learning complex, end-to-end manipulation skills through environmental interaction~\cite{kober2013reinforcement, mandlekar2020learning, wang2019efficient, zhu2018reinforcement, quillen2018deep, zhang2015towards}. However, widespread success is often hindered by challenges like sample efficiency and sparse rewards, which has prompted research into demonstration-guided learning~\cite{james2022q} and efficient architectures like graph-based RL frameworks~\cite{moghadam2023grasp}.

While end-to-end deep learning is a popular data-driven approach for robotic manipulation~\cite{fang2020learning, holladay2019force,lovchik1999robonaut, koppula2013learning}, the resulting "black-box" representations often lack the interpretability essential for generalizable control. Keypoint representations offer a structured alternative, providing a compact understanding of object geometry that improves generalization in contact-rich tasks~\cite{manuelli2019kpam, fang2020learning, park2008multiple }. However, the scalability of these methods has been hindered by a reliance on manual 3D annotation~\cite{manuelli2019kpam,holladay2019force}. Even recent automated 2D keypoint generation methods depend on traditional geometric techniques for the entire grasping process, which can overlook viable grasp points and restrict adaptability~\cite{sabzejou20232d}. This highlights a clear need for a more robust refinement strategy.

This work introduces a novel framework that bridges the gap between traditional geometric methods and data-driven learning for robotic grasping. The primary contribution of the present work is a Deep Q-Network (DQN) agent designed specifically for the task of refinement. Rather than learning to grasp, the proposed framework takes failed grasp candidates generated by a geometric algorithm from a 2D overhead image and iteratively adjusts their parameters—represented as \((x, y, \theta, w)\)—to optimize grasp stability and success rates.

\begin{figure}[t]
	\centering
	\includegraphics[width=0.8\columnwidth]{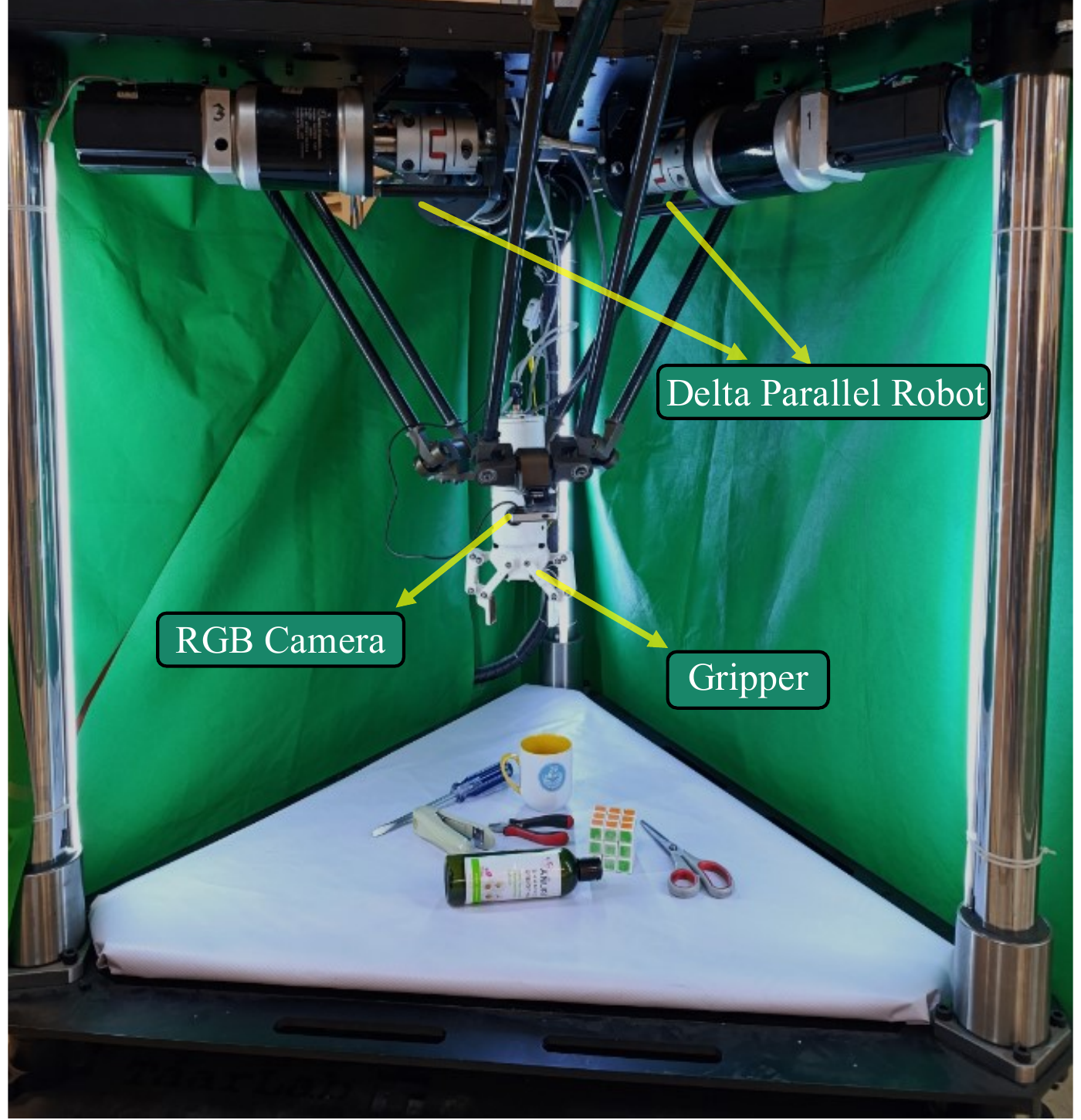} 
	\caption{The Delta parallel manipulator, equipped with a two-finger parallel-jaw gripper, is shown over the workspace. An overhead RGB camera captures the workspace.}
	\label{fig0}
\end{figure}

This work offers a distinct approach, complementing end-to-end grasp synthesis as a refinement module. By focusing the RL problem on local optimization around promising candidates rather than on global exploration, this method offers a more tractable, efficient, and scalable solution. The significance of this methodology lies in its demonstrated ability to systematically convert failed or suboptimal grasp attempts into successful, stable configurations. This enhances the overall robustness of automated grasping systems and presents a pragmatic pathway toward more adaptive and reliable robotic manipulation in complex tasks. The model's real-world performance was validated using a Delta parallel robot, shown in \hyperref[fig0]{Fig. 1}.

The remainder of this article is organized as follows. Section II explains the simulation setup, covering the simulation environment configuration, dataset characteristics and the methodology used for the initial labeling of objects through keypoint assignment. Section III outlines the methodology, including the grasp refinement framework, scoring system, state and action definitions, DQN implementation, and the training procedure. Section IV summarizes the findings and discusses them. Finally, conclusions are presented in Section~V.

\begin{figure*}[t]
	\centering
	\includegraphics[width=\textwidth]{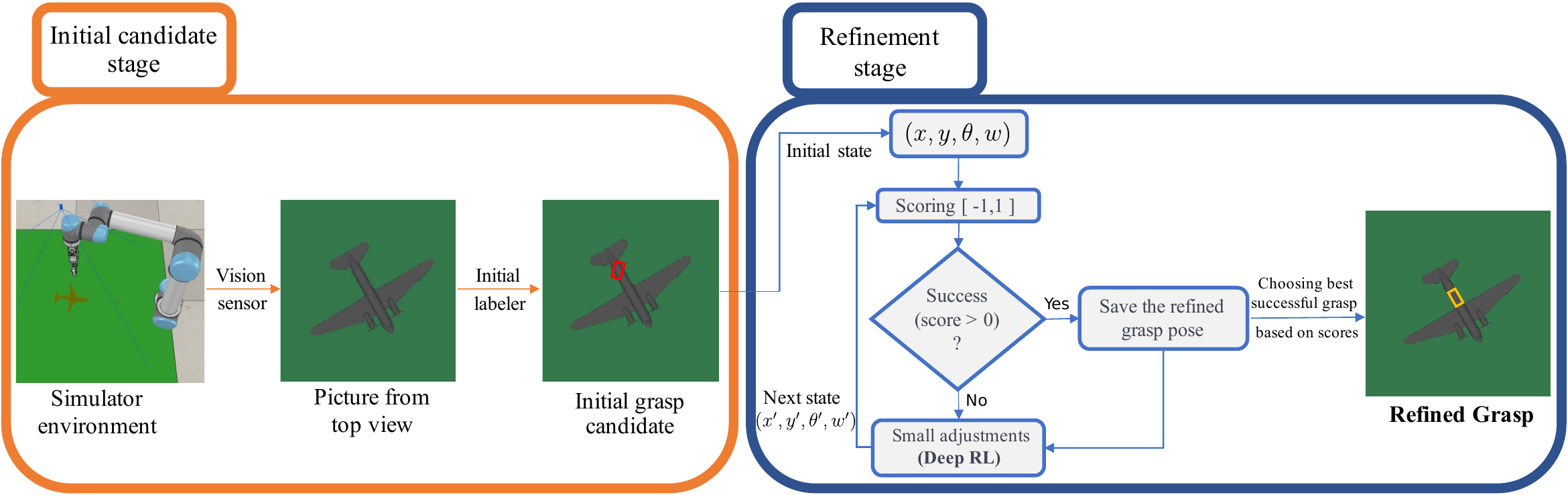} 
	\caption{The workflow of the refinement process. The initial grasp candidate, generated through geometric labeling, is refined within the simulation environment. Using reinforcement learning, the refinement stage iteratively adjusts the grasp pose \((x, y, \theta, w)\) through actions to maximize stability and minimize slippage, transforming a failed grasp into a successful one. The scoring mechanism, serving as the reward, evaluates each state and guides the agent toward optimal grasp configurations.}
	\label{fig1}
\end{figure*}

	\section{Dataset Generation and Labeling}
	This section elaborates on the simulation environment, the dataset used, how the simulation was employed to process the dataset, and the methodology of labeling (assigning key points) for objects. The simulation setup is described in detail below:
	
	\subsection{Simulation Environment and Configuration}
	The CoppeliaSim environment was employed for simulation due to its robust and feature-rich platform capabilities. This environment offers precise physics-based modeling and simulation, making it ideal for tasks like robotic grasping. A UR5 manipulator robot equipped with a RG2 gripper was used in the simulation. Additionally, an RGB-D vision sensor integrated into the simulation environment captured images of the virtual workspace from an overhead view, providing a comprehensive perspective for grasp analysis. The CoppeliaSim Python API facilitated real-time data exchange.

	\subsection{Object Dataset}
	The dataset used for this project was a random subset of the Dex-net dataset~\cite{7487342}, which contains 300 diverse object meshes. These meshes were imported into the CoppeliaSim environment for simulation. In order to facilitate the grasping process, a geometric-based algorithm, which was proposed by Sabzejou et al.~\cite{sabzejou20232d}, was utilized to initially label these objects by evaluating all possible grasp configurations. After the initial labels were obtained, they were tested in the simulator to observe whether they were successful grasps.
	
	Notably, 52 objects in the dataset did not exhibit any successful grasp configurations during the initial evaluation. From these, 14 objects were excluded from the refinement study as their geometries exceeded the physical limitations of the RG2 gripper (i.e., the maximum opening width). For the remaining 38 objects, one of the failed grasp candidates was randomly selected and subjected to the proposed refinement algorithm. This approach aimed to determine whether the algorithm could convert failed grasps into successful ones, demonstrating the robustness and adaptability of the proposed refinement method.
	
	\subsection{Initial Labeling}\label{AA}
	To initiate the labeling process, the objects were imported into the CoppeliaSim environment, and an RGB-D camera was used to capture overhead images of the scene. These images served as input for the previously mentioned geometric-based algorithm, proposed by Sabzejou et al.~\cite{sabzejou20232d}, which acted as an initial labeler. Labeling, in this context, refers to identifying initial grasp configurations for each object based on their geometrical shape.
	
	After obtaining the initial grasp candidates, they were tested in the simulation environment. The performance of each configuration was evaluated based on the quality of the grasp regarding stability and lift success. Scores were then assigned to each grasp configuration, providing a baseline for further refinement.

	\section{Grasp Refinement Framework}
	\subsection{Overview}
	The overall framework of the proposed system is illustrated in \hyperref[fig1]{Fig. 2}. The primary objective of the refinement method is converting failed grasps into successful ones and also enhancing the quality of already identified successful grasps regarding stability. The process begins with an initial grasp pose, which serves as input to the refinement model. The model systematically modifies the dimensions of the grasp pose and evaluates each variation in the simulation environment.
	
	Before the manipulator attempts to grasp an object, the position and orientation of the object are recorded. This initial state, referred to as the ``initial stage," provides a baseline for comparison. After the grasp attempt, the position and orientation of the object are recorded again, referred to as the ``final stage."\\Four outcomes are possible for each grasp pose:\\
	1) The manipulator fails to lift the object, instead merely pushing it across the surface or leaving it unmoved.\\
	2) The manipulator lifts the object, but due to excessive slippage caused by an improper grasp pose, the object falls off the gripper.\\
	3) The manipulator successfully lifts the object to an appropriate height (fully lift the object from the surface) with some rotation. \\
	4) The manipulator successfully lifts the object to an appropriate height with negligible slippage or instability.
	
	 After the manipulator attempts to grasp the object using a modified grasp pose, the simulator returns two outputs: a success flag indicating whether the grasp was successful and a score reflecting the stability of the grasp. By iteratively refining the grasp pose, the model aims to identify configurations that achieve the highest scores.
	
	\subsection{Scoring and Reward Structure}
	The scoring system evaluates grasp stability by measuring the amount of rotational slip (tilt), calculated during the grasp. Minimizing slippage is essential for achieving stable and secure grasps.
	
	Scoring is calculated based on the differences in the object's orientation between the initial and final stages. The simulator provides roll, pitch, and yaw values for both stages, which are used to compute the rotational slip $(|\Delta\theta|)$. Given that the maximum possible angular deviation is $\pi$ radians, the final score was normalized to a $[0, 1]$ range for successful grasps and to a $[-1, 0]$ range for failure grasps:
	\noindent  
	\begin{equation}
		\textbf{Score =} 
		\begin{cases}
			1 - \frac{|\Delta\theta|}{\pi} & \text{successful grasp} \\
			
			\\
			-1 + \frac{|\Delta\theta|}{\pi} & \text{failure grasp}
		\end{cases}
	\end{equation}

A perfectly stable, tilt-free grasp results in a score of~$+1$. 
Grasp failure is defined as the complete closure of the gripper in the final stage, indicating the absence of an object between its two fingers; this typically manifests as the object dropping during the lifting phase or the gripper being unable to lift the object from its initial position. 
For successful grasps (lift completed), smaller rotation differences between the initial and final stages indicate more stable and precise grasps, resulting in higher scores (approaching~$+1$). 
For failed grasps, the score reflects the extent to which the object was manipulated despite the failure. 
A completely missed grasp (no contact) yields~$-1$, while grasps that make contact and partially move or lift the object (even if it eventually falls) receive higher scores approaching~0. 
Thus, within the failed range~$[-1, 0)$, greater object motion or rotation indicates a grasp closer to success.

	\subsection{Actions and States Definition}
	The state in this work is defined by the grasp configuration, represented by a rectangle with four continuous parameters: the Cartesian coordinates of the rectangle's center $(x, y)$, its in-plane rotation $(\theta)$, and its width $(w)$. The state space is represented as a 4D vector $(x, y, \theta, w)$, which provides sufficient parameters for a two-finger robotic gripper.
	
	Actions are defined as discrete, small adjustments applied to the state parameters. The action space, represented as \((\Delta x, \Delta y, \Delta \theta, \Delta w)\), consists of incremental changes to the state variables which are selected from a discrete set of predefined small adjustments. These modifications allow the algorithm to explore and refine grasp configurations effectively. Additionally, a no-op action (no operation) is included, which leaves the state unchanged. This action is useful when the network determines that the current state is already optimal. The specific adjustments corresponding to each discrete action are detailed in Table~\ref{tab:action_space}.

	\begin{table}[t] 
	\centering
	\caption{Discrete Action Space for Grasp Refinement.}
	\label{tab:action_space}
	\begin{tabular}{@{}ccl@{}}
		\toprule
		\textbf{Action Index ($a$)} & \textbf{Adjustment $(\Delta x, \Delta y, \Delta\theta, \Delta w)$} & \textbf{Description} \\ \midrule
		
		0 & $(+1, 0, 0, 0)$ & Adjust X-pixel  \\
		1 & $(-1, 0, 0, 0)$ & Adjust X-pixel \\
		2 & $(0, +1, 0, 0)$ & Adjust Y-pixel  \\
		3 & $(0, -1, 0, 0)$ & Adjust Y-pixel  \\
		4 & $(0, 0, +2, 0)$ & Adjust rotation  \\
		5 & $(0, 0, -2, 0)$ & Adjust rotation  \\
		6 & $(0, 0, 0, +2)$ & Adjust width \\
		7 & $(0, 0, 0, -2)$ & Adjust width \\
		8 & $(0, 0, 0, 0)$ & No operation  \\ \bottomrule
	\end{tabular}
\end{table}

	\subsection{DQN Implementation and Training}
	The refinement algorithm is implemented using a Deep Q-Network (DQN)~\cite{Mnih2015_vb}, a technique that utilizes deep neural networks to approximate the Q-value function. A value-based DQN was selected as it is a sufficient and computationally efficient approach for the local refinement task, given the problem's low-dimensional state space.
	
	The DQN is trained to approximate the optimal Q-value function, $Q(s, a \mid \phi)$, where $\phi$ represents the network parameters. In this framework, the core components are defined as follows:
	
	\begin{itemize}
		\item State ($s$): The state $s_t = (x, y, \theta, w)$ is the 4D grasp configuration vector at time $t$.
		\item Action ($a$): The action $a_t$ is one of the 9 discrete adjustments $(\Delta x, \Delta y, \Delta \theta, \Delta w)$ selected from the predefined action space (Table~\ref{tab:action_space}).
		\item Reward ($r$): The immediate reward $r_t$ is the stability score, ranging from $[-1, +1]$, returned by the scoring mechanism. This score is calculated based on the object's rotational slip after a grasp attempt.
	\end{itemize}
	
	The network's parameters $\phi$ are optimized by sampling mini-batches of experiences $(s, a, r, s')$ from an experience replay buffer and minimizing the Mean Squared Error (MSE) loss:
	
	\begin{equation}
		L(\phi) = \mathbb{E}_{(s,a,r,s')} \left[ \left( Q(s, a \mid \phi) - y \right)^2 \right],
		\label{eq:loss} %
	\end{equation}
	
	where the target y is defined as:
	
	\begin{equation}
		y = r + \gamma \max_{a'} Q(s', a' \mid \phi),
	\end{equation}
	
	Here, $\gamma$ is the discount factor, which was set to 0.9.
	
	The Q-network is a fully connected neural network with three hidden layers (256, 128, and 64 neurons, respectively), each using a ReLU activation function and Layer Normalization. The network was trained using the Adam optimizer with a learning rate of 0.005. For exploration, an $\epsilon$-greedy policy was employed; the exploration rate $\epsilon$ was initially set to 1.0 and decayed by a factor of 0.999 per step to a minimum of 0.01. The agent was trained for $E = 25$ episodes for each object, with each episode consisting of a maximum of $T = 120$ steps. The agent's experience replay buffer was sampled using a batch size of 32. This process enables the agent to iteratively refine the grasp configuration by applying small displacement actions $(\Delta x, \Delta y, \Delta \theta, \Delta w)$ to the current pose. The learning process, which refines an initial grasp candidate, is summarized in Algorithm~\ref{alg:refinement}.

\begin{algorithm}[h!]
	\caption{Grasp Refinement using Deep Q-Learning }
	\label{alg:refinement}
	\textbf{Input:} Initial failed grasp $s_{\text{init}}=(x,y,\theta,w)$, action set $A$ (\textbf{Table~\ref{tab:action_space}}), Number of episodes $E$, max steps per episode $T$ \\
	\textbf{Output:} The refined grasp configuration $s_{best}$
	\begin{algorithmic}[1]
		\State Initialize Q-network $Q_{\phi}$ with random weights
		\State Initialize target network $Q_{\phi'}$ with weights $\phi' \leftarrow \phi$
		\State Initialize replay buffer $\mathcal{D}$
		\State Initialize exploration rate \( \epsilon \)
		\State Initialize $s_{best} \leftarrow s_{init}$ and $score_{best} \leftarrow -1.0$
		
		\For{$episode = 1$ \textbf{to} $E$}
		\State $s_t \leftarrow s_{init}$
		\For{$t = 1$ \textbf{to} $T$}
        \State Select $a_t$ using $\epsilon$-greedy policy based on $Q_{\phi}(s_t)$
	\State Calculate next state $s_{t+1}$ by applying $a_t$ to $s_t$
\State Execute grasp $s_{t+1}$ in simulator $N$ times 
\State $r_t \leftarrow \text{AverageScore}(s_{t+1})$ 
		\State Store transition $(s_t, a_t, r_t, s_{t+1})$ in $\mathcal{D}$
		
		\If{$r_t > score_{best}$}
		\State $score_{best} \leftarrow r_t$
		\State $s_{best} \leftarrow s_{t+1}$
		\EndIf
		
		\State $s_t \leftarrow s_{t+1}$
		
		\If{$|\mathcal{D}|$ is large enough}
		\State Sample a random minibatch from $\mathcal{D}$
		\State Perform a gradient descent step on the loss
		\Statex \hspace*{4.5em}function (from \textbf{Eq.~\eqref{eq:loss}}) to update $\phi$
		\EndIf
       \If{$r_t = 1$}
        \State \textbf{break}
        \EndIf
		\EndFor
		\State Decay exploration rate $\epsilon$
        \State Update target network: $\phi' \leftarrow \phi$
		\EndFor
		\State \textbf{return} $s_{best}$
	\end{algorithmic}
\end{algorithm}

	\begin{figure}[t]
	\centerline{\includegraphics[width=1\columnwidth]{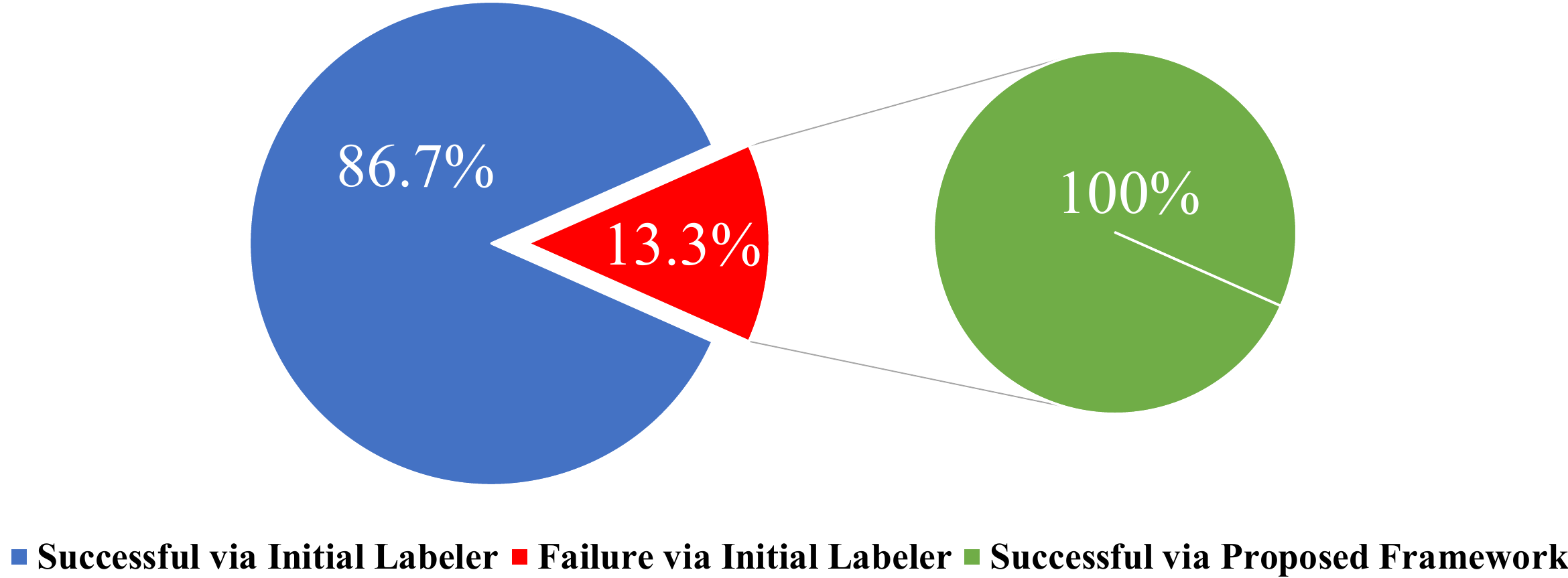}}
	\caption{Outcomes of challenging objects.}
	\label{fig2}
	\end{figure}
  \begin{figure*}[t]
    \centering
    \begin{minipage}{\columnwidth} 
        \centering
        \includegraphics[width=0.9\linewidth]{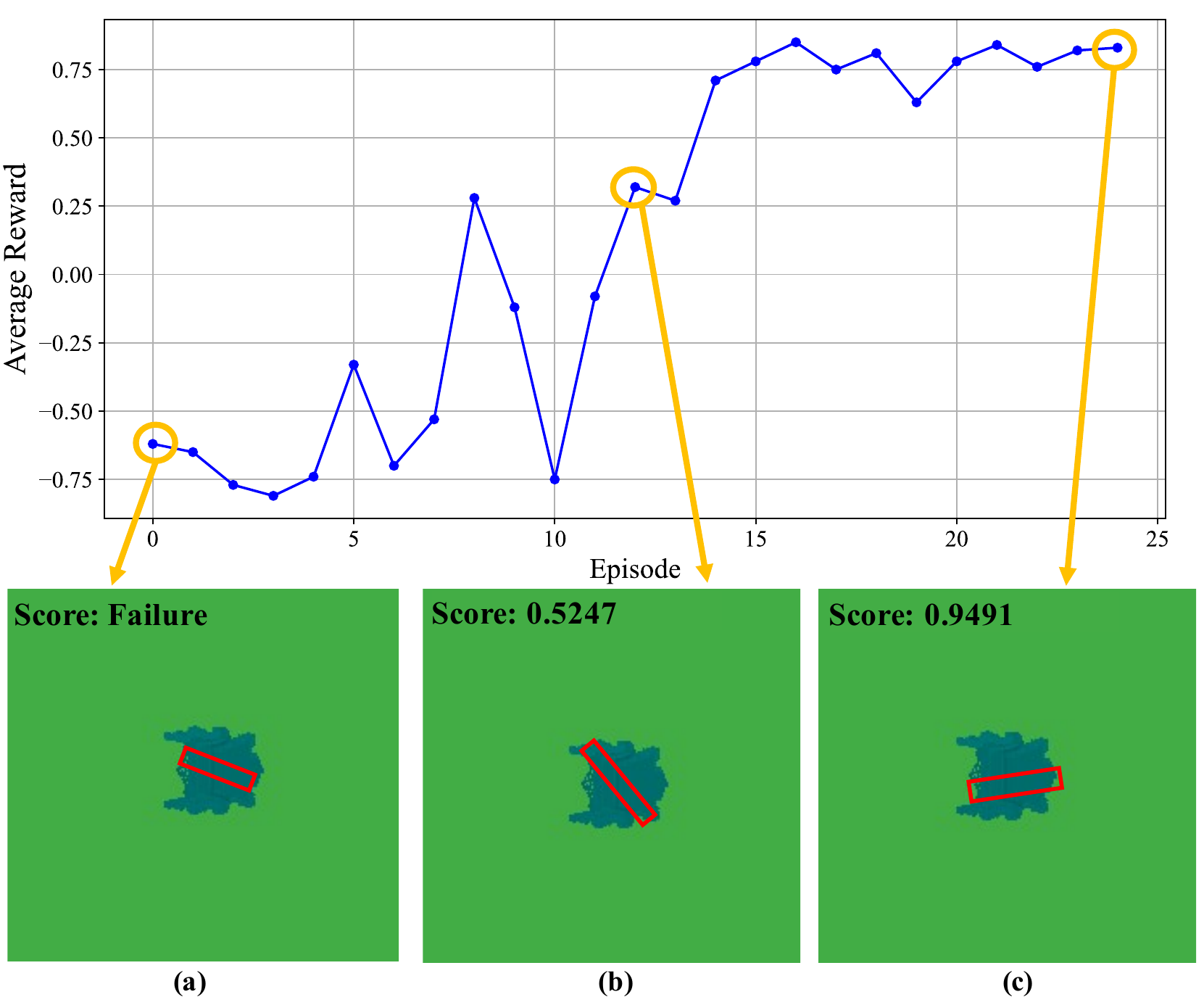}
        \caption{Average reward per episode and the corresponding grasp configuration.}
        \label{fig3}
    \end{minipage}
    \hfill 
    \begin{minipage}{\columnwidth}
        \centering
        \includegraphics[width=0.9\linewidth]{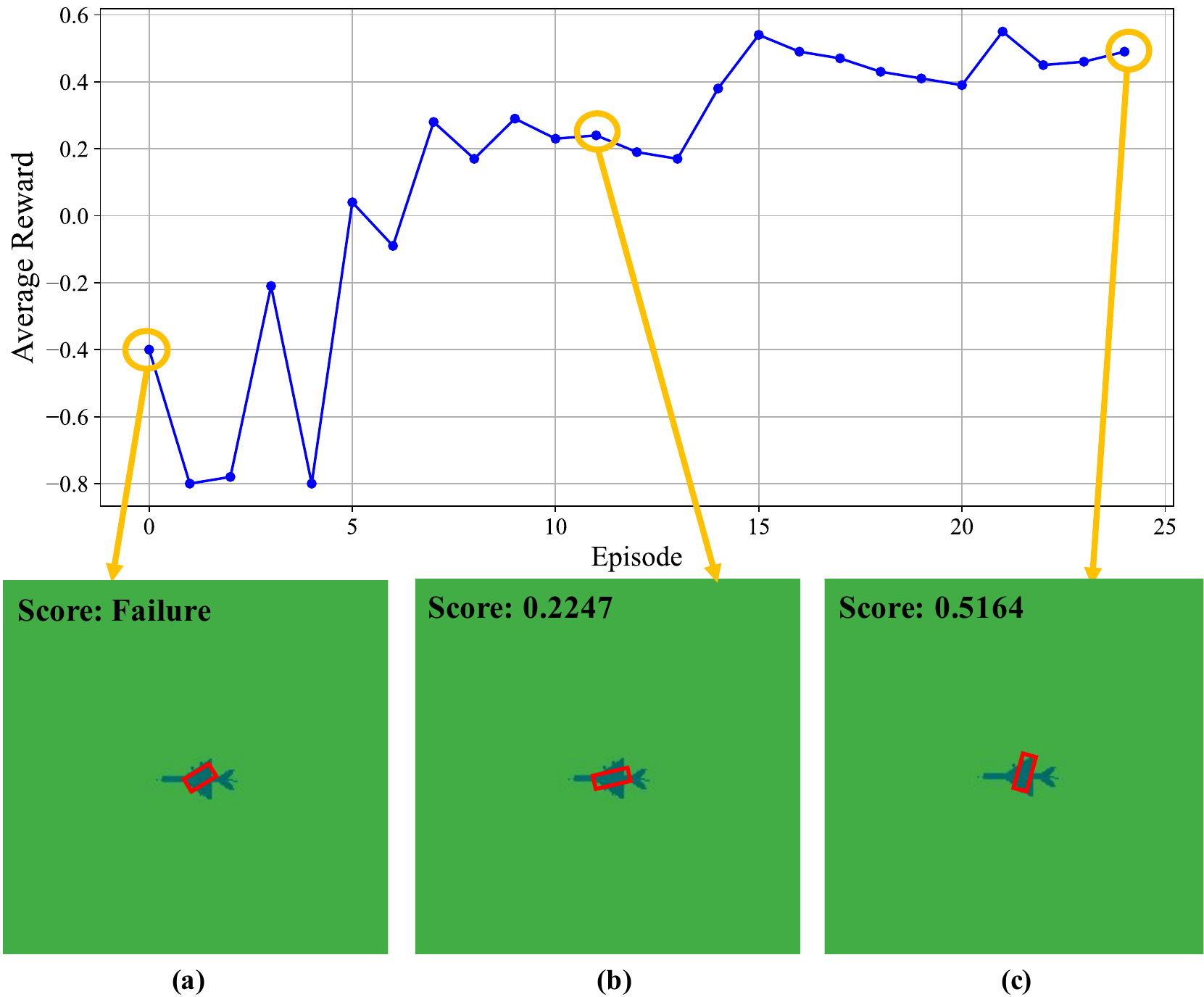}
        \caption{Average reward per episode and the corresponding grasp configuration.}
        \label{fig4}
    \end{minipage}
\end{figure*}
	\section{Results and Discussion}
	The proposed framework underwent a rigorous training procedure to evaluate its performance. A random subset of 300 objects from the Dex-Net dataset~\cite{7487342} was used, and 2D top-view images of these objects were labeled using the geometric-based algorithm proposed by Sabzejou et al.~\cite{sabzejou20232d}. The algorithm processes each object separately, and the resulting labeled grasp configurations were subsequently tested in a simulation environment. Among the 300 objects, 14 of them could not be grasped due to physical limitations, such as the maximum opening width of the gripper, so they were excluded from the analysis. From the remaining 286 objects, 38 objects were found to lack any grasp configuration that allowed the robot to grasp and lift them to an appropriate height.
	
	The proposed refinement framework was applied to these 38 objects. For all of these 38 objects, successful refined grasps were identified. Thus, the success rate of the proposed framework was 100\% (\hyperref[fig2]{Fig. 3}). To ensure robustness, each grasp configuration was executed three times and deemed successful only if all trials succeeded; otherwise, the state was labeled a failure. This repeated evaluation confirms that the resulting grasp configurations are reliable and stable.

	The effectiveness of the DQN training process is demonstrated through \hyperref[fig3]{Fig. 4} and \hyperref[fig4]{5}, which display the agent’s average reward per episode and illustrate its progressive improvement in grasp configurations.

		\begin{figure*}[t] 
		\centerline{\includegraphics[width=2\columnwidth]{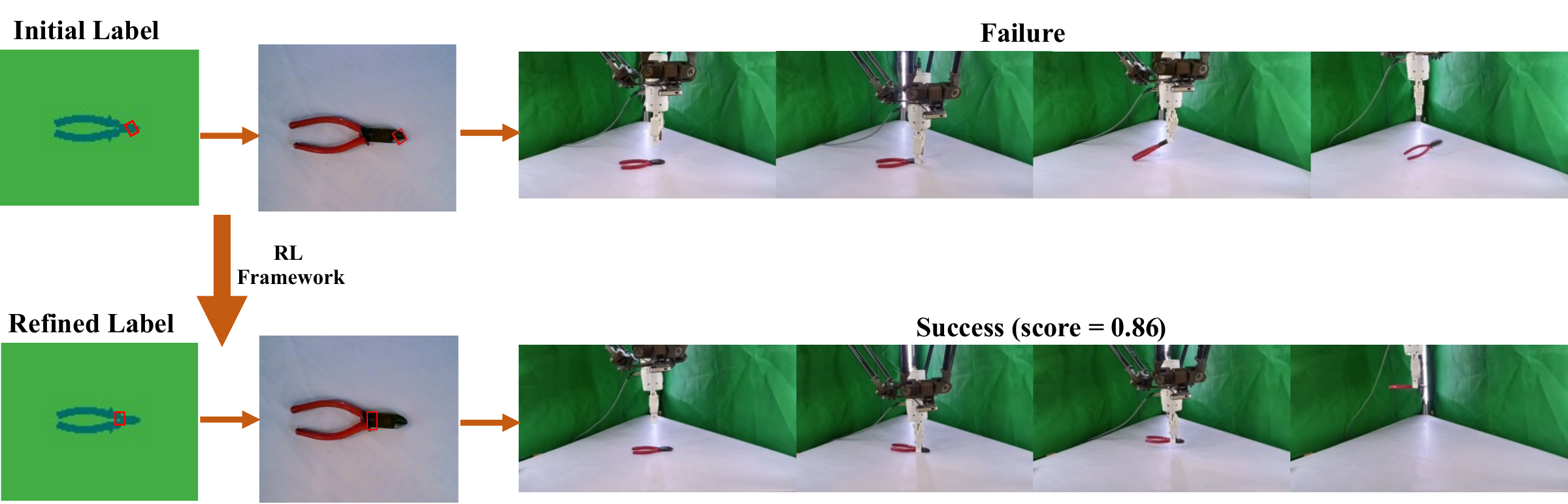}} 
		\caption{Sim-to-real validation of the proposed grasp refinement framework using a pair of pliers. 
			\textbf{(Top)} An initial grasp candidate, generated by the geometric algorithm, proves unstable, \textbf{leading to failure in both the simulation and the real-world execution}. 
			\textbf{(Bottom)} After processing the failed candidate, the proposed RL framework generates a refined grasp pose. This new pose, \textbf{which was identified as successful in simulation}, enables a stable lift with a Delta parallel robot, confirming the direct transfer of the refined grasp's effectiveness from simulation to a real-world task.}
	
		\label{fig5} 
	\end{figure*}

	\hyperref[fig3]{Fig. 4(a)} and \hyperref[fig4]{Fig. 5(a)} show the initial unsuccessful candidates. Initial attempts yielded minimal rewards, reflecting the agent's lack of knowledge. As training progressed, the average reward obtained by the agent increased steadily, as shown in \hyperref[fig3]{Fig. 4(b)} and \hyperref[fig4]{Fig. 5(b)}. At this stage, the grasp configurations began to show minor improvements, albeit with some residual similarities to the initial failed configurations.
	
	\hyperref[fig3]{Fig. 4(c)} and \hyperref[fig4]{5(c)} illustrate the  grasp configurations from episodes with the highest average rewards. These configurations represent the refined grasps identified by the proposed framework. The noticeable differences from initial configurations highlight the effectiveness of the DQN algorithm in optimizing grasp poses. These increased rewards correspond to stable grasps meeting the success criteria.

	The evolution of average rewards in \hyperref[fig3]{Fig. 4} and \hyperref[fig4]{5} underscores the framework's learning capabilities, affirming the robustness and adaptability of the proposed refinement framework in improving the quality and success rate of grasping tasks. The progressive refinement of grasp configurations demonstrates the potential of reinforcement learning to address challenges in robotic grasping, particularly for objects initially deemed ungraspable by geometric algorithms.
	Finally, the proposed framework was evaluated on a physical testbed to verify the transferability of refined grasps from simulation to the real world. An object from the dataset, whose grasp configuration had been labeled as a failure in simulation, was selected; a corresponding physical real object was identified and assigned the same grasp label. The initial (failed) grasp was executed using the Delta parallel manipulator and likewise failed in the real experiment. The failed grasp configuration was then processed by the refinement algorithm, which produced a refined grasp that was successful in simulation. This refined grasp was transferred to the physical object and executed with the manipulator, resulting in a successful and stable lift. An example of this sim-to-real validation is presented in \hyperref[fig5]{Fig. 6}.
	
	\section{CONCLUSION}
	This work introduces a novel reinforcement learning-based framework for refining robotic grasp configurations, combining keypoint-based object representations with a Deep Q-Network (DQN). By leveraging a geometric-based labeling algorithm as a baseline, the framework iteratively optimizes grasp poses, transforming initially failed grasps into stable and precise configurations. Simulation results demonstrate the framework's effectiveness in addressing limitations of conventional labeling methods, with significant improvements in grasp success rates for challenging objects.
	
	From the 52 objects initially labeled ungraspable, 38 fell within hardware limits and the proposed refinement framework produced successful refined grasps for all 38 (100\% success). These findings underscore the framework's ability to systematically explore and refine grasp configurations, even for objects that pose significant challenges due to their geometry or orientation. The progressive increase in rewards validates the DQN's ability to transform low-reward failures into optimized configurations. This process not only converts failed grasps into successful ones but also enhances the stability of existing candidates.
	Sim-to-real validations were performed: failed simulated grasp was refined by the proposed algorithm, and the refined configuration yielded a successful lift on the corresponding physical object, confirming the practical effectiveness of the refinement method. Future work will focus on enabling the DQN to generalize better to unseen data. By enhancing its adaptability to diverse and unpredictable scenarios, this research contributes toward the development of more scalable and versatile robotic systems capable of robust manipulation in complex, real-world environments.

	\bibliographystyle{IEEEtran}
	\bibliography{references}
	
\end{document}